\documentclass[runningheads]{llncs}

\usepackage[T1]{fontenc}
\usepackage{graphicx}
\usepackage{booktabs}
\usepackage{multirow}
\usepackage{colortbl}
\usepackage{color}
\usepackage[pagebackref,breaklinks,colorlinks]{hyperref}

\usepackage{wrapfig}
\usepackage{orcidlink}
\usepackage{xr-hyper}
\usepackage{algorithm}
\usepackage{algpseudocode}
\usepackage{amsmath,bm}
\usepackage{bbm}
\usepackage{amsfonts}
\usepackage{caption}
\usepackage{subfig}
\usepackage{siunitx}

\usepackage{graphicx}
\usepackage[table]{xcolor} 
\usepackage{color} 
\definecolor{Gray}{gray}{0.9}
\definecolor{ours_color}{rgb}{0.92, 0.97, 1}
\newcommand{\grayc}{\cellcolor{Gray}}

\usepackage[euler]{textgreek}
\usepackage{arydshln}
\usepackage{marvosym}

\usepackage{tikz}

\makeatletter
\def\adl@drawiv#1#2#3{%
        \hskip.5\tabcolsep
        \xleaders#3{#2.5\@tempdimb #1{1}#2.5\@tempdimb}%
                #2\z@ plus1fil minus1fil\relax
        \hskip.5\tabcolsep}
\newcommand{\cdashlinelr}[1]{%
  \noalign{
           \global\let\@dashdrawstore\adl@draw
           \global\let\adl@draw\adl@drawiv}
  \cdashline{#1}
  \noalign{\global\let\adl@draw\@dashdrawstore
           }}
\makeatother

\begin{document}

\title{Trustworthy Endoscopic Super-Resolution}
\titlerunning{Trustworthy Endoscopic Super-Resolution}
 

%
%

\author{Julio Silva-Rodríguez \and Ender Konukoglu}
\authorrunning{J. Silva-Rodr\'iguez and E. Konukoglu}
\institute{Computer Vision Lab, ETH Zurich, Zurich, Switzerland \\
    \email{jusilva@ethz.ch}}


\maketitle          

\begin{abstract}

Super-resolution (SR) models are attracting growing interest for enhancing minimally invasive surgery and diagnostic videos under hardware constraints. However, valid concerns remain regarding the introduction of hallucinated structures and amplified noise, limiting their reliability in safety-critical settings. We propose a direct and practical framework to make SR systems more trustworthy by identifying where reconstructions are likely to fail. Our approach integrates a lightweight error-prediction network that operates on intermediate representations to estimate pixel-wise reconstruction error. The module is computationally efficient and low-latency, making it suitable for real-time deployment. We convert these predictions into operational failure decisions by constructing Conformal Failure Masks (CFM), which localize regions where the SR output should not be trusted. Built on conformal risk control principles, our method provides theoretical guarantees for controlling both the tolerated error limit and the miscoverage in detected failures. We evaluate our approach on image and video SR, demonstrating its effectiveness in detecting unreliable reconstructions in endoscopic and robotic surgery settings. To our knowledge, this is the first study to provide a model-agnostic, theoretically grounded approach to improving the safety of real-time endoscopic image SR. Code is available: \href{https://github.com/jusiro/Endoscopic-CFM}{Endoscopic-CFM}.

\end{abstract}

\section{Introduction}
\label{sec:intro}

Minimally invasive surgery and diagnostic procedures, including laparoscopic and robot-assisted techniques, have advanced surgical practice by reducing recovery time, tissue trauma, and scarring~\cite{mack2001minimally}. In this context, high-quality imaging is essential for surgical safety, precise dissection, and identification of critical anatomical structures~\cite{goh2010minimally}. However, acquiring high-resolution images in clinical practice remains challenging due to hardware constraints, artificial noise sources, and patient comfort requirements, particularly in real-time clinical settings~\cite {chow2016review}.

Recent advances in machine learning aim to overcome these limitations with super-resolution (SR) deep neural networks that recover fine details and reduce noise. Lightweight models for real-time single-image~\cite{liang2021swinir,niu2020single} and video~\cite{chan2021basicvsr,chan2022basicvsr_pp} SR have shown promising results, including for endoscopic data~\cite{liu2025medvsr}. Nevertheless, SR models, and generative models in general, raise concerns about reliability, potentially producing unnoticed hallucinations and amplifying artifacts. In this context, medical SR approaches must provide not only accurate reconstructions but also robust estimates of failure awareness to ensure safe clinical use.

\noindent\textbf{Pitfalls of current uncertainty quantification frameworks.} Failure detection has often been approached through the lens of uncertainty estimation, under the assumption that uncertain predictions correlate with high error. In image classification, softmax scores often correlate with correct class predictions~\cite{hendrycks17baseline} and can be further calibrated~\cite{temp_scaling}. However, these approaches do not directly extend to continuous regression outputs, as in image SR. Other alternatives leverage stochastic processes (e.g., Monte Carlo dropout~\cite{mcd} or diffusion-based models~\cite{adame2025image}) or ensembles~\cite{lakshminarayanan2017simple} to estimate predictive variability as a proxy for uncertainty. While effective, such methods typically require multiple forward passes or architectural modifications, making them less suitable for real-time pipelines. A lighter alternative is to learn failure estimates directly with auxiliary networks. These approaches have been explored mainly in discriminative contexts, such as classification and segmentation~\cite{corbiere2019addressing,auxiliary_models} or multi-modal understanding~\cite{lafon2025vilu}, where reliability can be tied to well-defined probabilistic scores, e.g., true-class probabilities. Extending these ideas to image SR is less straightforward. In SR, outputs are continuous and structured, and failure must be defined relative to task-specific thresholds rather than an intrinsic notion of correctness. Crucially, beyond these limitations, none of these alternatives alone yield operational failure decisions with robust, distribution-free guarantees suitable for clinical use.

\noindent\textbf{Conformal prediction} (CP) ~\cite{learning_ny_transduction,vovk_book} is a framework that aims to provide model-agnostic, distribution-free, and theoretically valid guarantees on machine learning models' outputs. In its most widely adopted version, split conformal prediction~\cite{inductive_ci,vovk_book}, a calibration subset is used to post-process neural network outputs into operational decisions, e.g., decision intervals for regression~\cite{romano2019conformalized,stankeviciute2021conformal,i2iregression} or class sets for classification~\cite{lac,raps}, that guarantee a given coverage of correctness. Despite growing interest, CP remains relatively underexplored in medical image computing, although recent work has shown promising results in image classification~\cite{aiii_conf_mic,Che_Modeling_MICCAI2024}, grading~\cite{sev_grading_conf}, segmentation~\cite{kandinskycp2023,bereskasacp,Mossina_2025_conformal_morpho}, multi-modal models~\cite{fca25,scat25}, and surgical instrument forecasting~\cite{SanSar_Conformal_MICCAI2025}, among other tasks~\cite{SiWen_Reliable_MICCAI2025,TenJac_Conformal_MICCAI2025,cheung2024metric,Lam_Robust_MICCAI2024}. In the computer vision community, CP for image-to-image generation has been typically addressed by producing interval-valued outputs~\cite{i2iregression,teneggi2023trust} instead of single images. In such a pipeline, the reliability is communicated qualitatively to the practitioner via the interval size; e.g., pixels with larger intervals are likely to be less reliable. However, as recently shown in~\cite{adame2025image}, these heatmaps are hardly interpretable and do not yield fully operational failure decisions. In contrast, the very recent work in~\cite{adame2025image}, which is more closely aligned with our efforts, explores the creation of binary masks that explicitly highlight image reconstruction errors.

\begin{figure*}[t!]
    \begin{center}
        \setlength{\tabcolsep}{0.5pt}
        \renewcommand{\arraystretch}{0.5}
        \begin{tabular}{c}

            \setlength{\tabcolsep}{0.5pt}
            \renewcommand{\arraystretch}{0.5}
            \hspace{-1.3mm}
            \begin{tabular}{ccc}

                \begin{tabular}{c}
                    \includegraphics[width=0.1\linewidth]{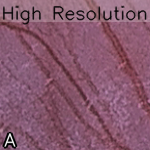}  \\
                    \includegraphics[width=0.1\linewidth]{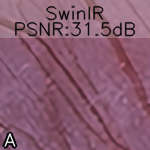} \\
                    \includegraphics[width=0.1\linewidth]{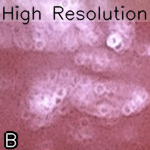}  \\
                    \includegraphics[width=0.1\linewidth]{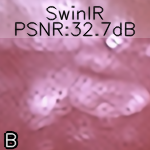} \\
                \end{tabular} 
                &
                \begin{tabular}{c}
                    \includegraphics[width=0.74\linewidth]{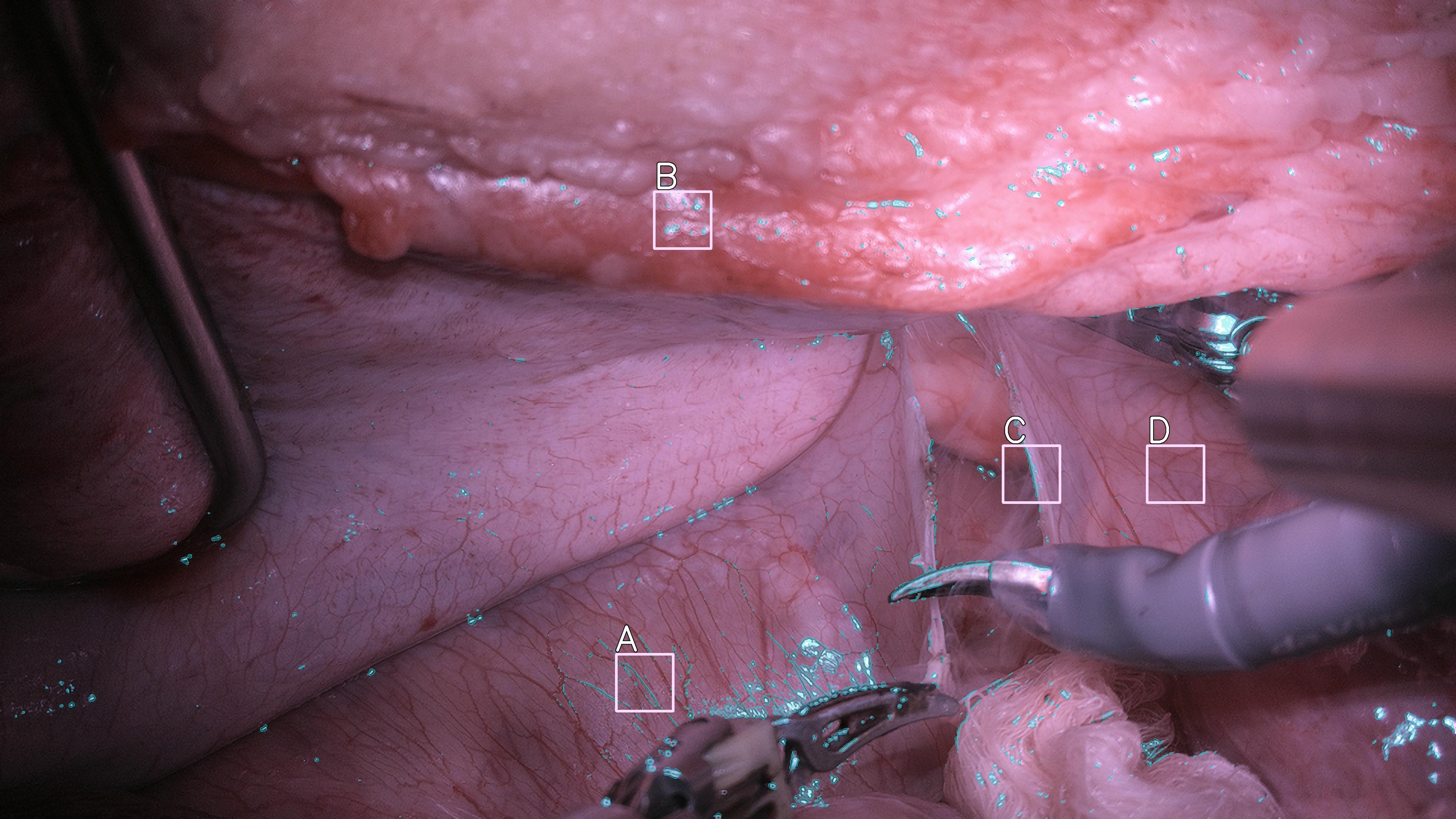} \\
                \end{tabular} 
                &
                \begin{tabular}{c}
                    \includegraphics[width=0.1\linewidth]{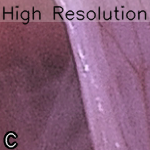}  \\
                    \includegraphics[width=0.1\linewidth]{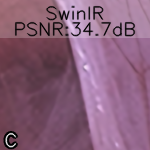} \\
                    \includegraphics[width=0.1\linewidth]{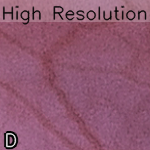}  \\
                    \includegraphics[width=0.1\linewidth]{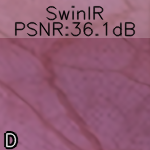} \\  
                \end{tabular} 
                \\

            \end{tabular} \\

            \hspace{-0.1mm}
            \setlength{\tabcolsep}{0.5pt}
            \renewcommand{\arraystretch}{0.5}
            \begin{tabular}{cccc}
                \includegraphics[width=0.255\linewidth]{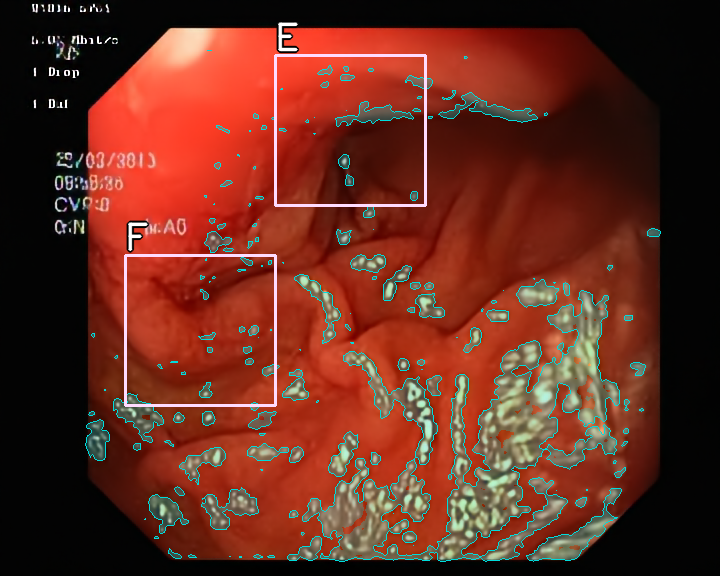} 
                &
                \raisebox{11.5mm}{
                \begin{tabular}{cc}
                    \includegraphics[width=0.1\linewidth]{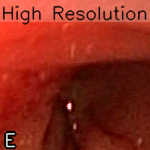} &
                    \includegraphics[width=0.1\linewidth]{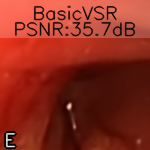} \\
                    \includegraphics[width=0.1\linewidth]{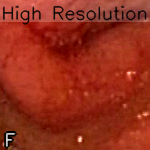} &
                    \includegraphics[width=0.1\linewidth]{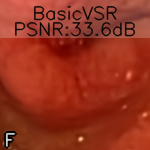} \\     
                \end{tabular} 
                }
                &
                \includegraphics[width=0.255\linewidth]{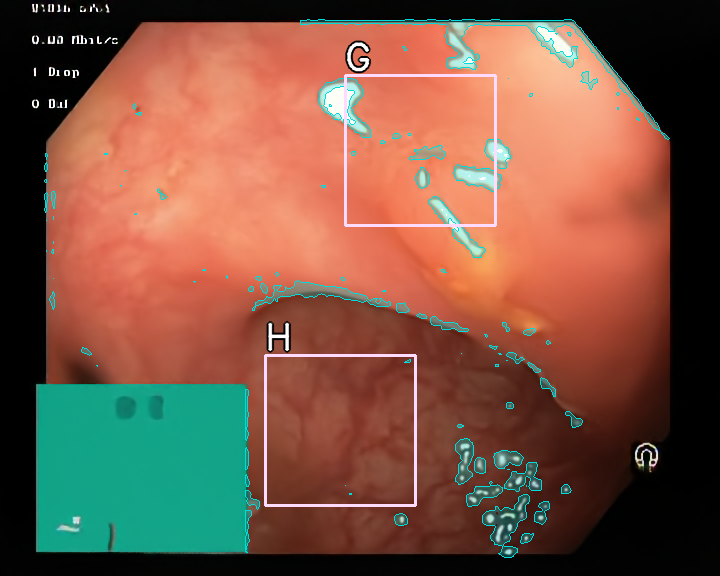}
                &
                \raisebox{11.5mm}{
                \begin{tabular}{cc}
                    \includegraphics[width=0.1\linewidth]{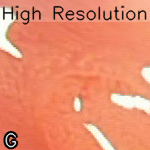} &
                    \includegraphics[width=0.1\linewidth]{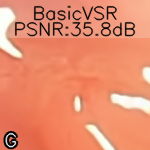} \\
                    \includegraphics[width=0.1\linewidth]{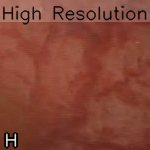} &
                    \includegraphics[width=0.1\linewidth]{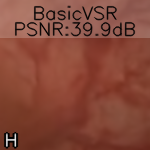} \\     
                \end{tabular} 
                }
                \\
            \end{tabular}

        \end{tabular}
    \caption{\textbf{Examples of our Conformal Failure Masks for trustworthy SR.}}
    \label{fig:qualitative}
    \end{center}
\end{figure*}

\noindent\textbf{Contributions.} In this work, we address the challenge of quantifying the reliability of SR models, focusing on an application in endoscopic imaging. Specifically, we propose a novel bi-level conformal risk control procedure that allows the user to control both the permissible level of reconstruction error and miscoverage on detected failures. Technically, we first propose learning a pixel-level error estimate using intermediate representations from a frozen SR model, with a lightweight module, termed the \emph{Reconstruction Error Network}. We then propose creating operational \emph{Conformal Failure Masks} (CFMs, \textit{see} Fig.~\ref{fig:qualitative}, \textit{blue-ish masks}) to identify generated regions that practitioners should not trust. For this purpose, we introduce the concept of \emph{reconstruction failure levels} to create binary failure-detection sub-tasks (e.g., as measured by PSNR). Our CFMs are then created by rejecting pixels whose estimated errors exceed a threshold. Specifically, the operative rejection threshold is chosen to control the false-negative rate at the target failure level by building on well-established theoretical guarantees for conformal risk control~\cite{angelopoulos2024conformal}. Finally, we present extensive experiments demonstrating that our framework can successfully measure and control the reconstruction error across three tasks and SR models. By controlling both the desired failure level and miscoverage, practitioners can flexibly generate practical reliability masks to support more informed clinical decisions and greater trust in technology. 

\section{Methods}
\label{sec:background}

\noindent\textbf{{Image SR}.} Let $\mathcal{X} \subset \mathbb{R}^{\Omega \times C}$ denote the low-resolution (LR) image space, where $\Omega \subset \mathbb{Z}^2$ is the discrete spatial domain, $\Omega=\{0, \dots, W-1\} \times \{0, \dots, H-1\}$, and $C$ denotes the number of channels. We define the high-resolution (HR) image space $\mathcal{Y} \subset \mathbb{R}^{\Omega' \times C}$, where, given an upscaling factor $s \in \mathbb{N}, s>1$, the HR spatial domain is $\Omega' = \{0, \dots, sW-1\} \times \{0, \dots, sH-1\}$, so that $|\Omega'| = s^2 |\Omega|$. Super-resolution (SR) aims to recover an HR image from its LR observation by learning a mapping $f_\theta : \mathcal{X} \rightarrow \mathcal{Y}$, parameterized by a deep neural network with parameters $\theta = \{\theta_{\mathrm{enc}}, \theta_{\mathrm{rec}}\}$. The network is typically composed of two stages: a feature extraction (encoder/bottleneck) module $f_{\theta_{\mathrm{enc}}} : \mathcal{X} \rightarrow \mathcal{F}$, which maps the input image into a deep latent feature space $\mathcal{F} \subset \mathbb{R}^{\Omega \times F}$ with $F$ feature channels, and a reconstruction (upsampling) module $f_{\theta_{\mathrm{rec}}} : \mathcal{F} \rightarrow \mathcal{Y}$, which performs spatial upsampling followed by nonlinear transformations to reconstruct the HR image.

\subsection{Lightweight SR error quantification}
\label{ssec:failure_network}

We first aim to obtain a measure of error for pixel-level predictions given a pre-trained SR model. For that purpose, we propose a data-driven learning procedure in which an auxiliary module projects the intermediate feature representations of the SR encoder into a predictive unreliability measure, i.e., 
$f_\Phi : \mathcal{F} \rightarrow \hat{\mathcal{E}}$, where $\hat{\mathcal{E}} \subset \mathbb{R}^{\Omega' \times 1}$. 
The function $f_\Phi$, parameterized by a deep neural network $\Phi$ and coined the \emph{Reconstruction Error Network}, is fitted to produce a continuous response, with larger values expected to align with higher reconstruction error.

\noindent\textbf{Optimization.} 
Let $\mathcal{D}_{\mathrm{train}} = \{ (\mathbf{X}_i, \mathbf{Y}_i) \}_{i \in \mathcal{T}}$, with $\mathcal{T} = \{ i \in \mathbb{N} : 0 \le i \le I-1 \}$, such that $\mathbf{X}_i \in \mathcal{X}$ and $\mathbf{Y}_i \in \mathcal{Y}$ are paired LR and HR images, denote a small training dataset with fresh data points, not used to pre-train the SR model. 
Also, let $\hat{\mathbf{Y}}_i = f_\theta(\mathbf{X}_i)$ denote the reconstructed image and $\mathbf{F}_i = f_{\theta_{\mathrm{enc}}}(\mathbf{X}_i)$ its intermediate feature representation. The network is optimized via mini-batch gradient descent to approximate the pixel-level squared error between predicted and reference HR images from the intermediate features:
\begin{align}
\label{eq:train_failurenet}
\min_{\Phi} \ \frac{1}{|\mathcal{T}| |\Omega'|} 
\sum_{i \in \mathcal{T}} 
\sum_{\Omega'} \Big( \frac{1}{C} \sum_{c=0}^{C-1} (\hat{\mathbf{Y}}_i - \mathbf{Y}_i)^2 - f_\Phi(\mathbf{F}_i) \Big)^2.
\end{align}
\noindent\textbf{Architecture.} 
Bicubic interpolation is used for initial upsampling, followed by a sequence of convolutional blocks and a final linear layer. Each block contains standard operations, i.e., convolution, ReLU, and batch normalization. This simple design provides a lightweight solution that can run in parallel with the SR model's reconstruction block without introducing additional latency.

\subsection{Conformal Failure Masks}
\label{ssec:cfm}

We now propose transforming the pixel-level continuous error scores into operational binary masks, referred to as \emph{Conformal Failure Masks} (CFMs). 

\noindent\textbf{Reconstruction failure levels.} Unlike standard classification tasks, errors in image reconstruction are continuous, reflecting varying degrees of incorrectness depending on the difficulty of the task. To enable binary decisions (failure / non-failure) from continuous errors, we first define a per-pixel observed error map $\mathbf{E}_i \in \mathbb{R}^{\Omega' \times 1}$ between the predicted and reference HR images, e.g., using peak signal-to-noise ratio (PSNR) or squared difference. In practice, we adopt PSNR, as is standard in SR~\cite{chan2021basicvsr,chan2022basicvsr_pp,liu2025medvsr,adame2025image}. Then, for a given failure level $\tau^{\mathrm{fail}} \ge 0$ (e.g., PSNR = 22 dB), the target binary mask is defined as $\mathbf{M}_i^{\tau^{\mathrm{fail}}} = \mathbf{1}\{\mathbf{E}_i^{\mathrm{PSNR}} \le \tau^{\mathrm{fail}}\} \in \{0,1\}^{\Omega' \times 1}$, where $\mathbf{1}\{\cdot\}$ is the indicator function. This procedure converts reconstruction errors into a set of binary classification sub-tasks, each corresponding to a user-controlled specific failure level, $\tau^{\mathrm{fail}}$.

\noindent\textbf{Bi-level risk control.} 
Next, given the estimate of our error network for a particular image, $\hat{\mathbf{E}}_i \in \hat{\mathcal{E}}$, these scores are binarized using an operative threshold, $\tau$, to approximate its oracle failure mask at a given failure level, $\mathbf{M}_i^{\tau^{\mathrm{fail}}}$, such that $\hat{\mathbf{M}}_i^{\tau} = \mathbf{1}\{\hat{\mathbf{E}}_i \ge \tau\} \in \{0,1\}^{\Omega' \times 1}$. However, due to imperfect error estimation, the operative threshold must balance a suitable risk measure. In clinical settings, a user may wish to control the number of false negatives (SR hallucinations or noisy regions that are not flagged), that is, the false-negative-rate (FNR), so that it does not exceed a predefined frequency, $\alpha \in [0,1]$. To do so, we propose a conformal procedure to establish the operative threshold that theoretically guarantees a miscoverage rate ($\alpha$) for an arbitrary failure level ($\tau^{\mathrm{fail}}$):
\begin{align}
\label{guarantee}
\mathrm{FNR}(\tau^{\mathrm{fail}}, \tilde{\tau}) \le \alpha.
\end{align}
\noindent Let us assume access to a set of LR and HR images different from those used to train the SR and error networks (the calibration set in split conformal prediction), $\mathcal{D}_{\mathrm{cal}} = \{ (\mathbf{X}_i, \mathbf{Y}_i) \}_{i \in \mathcal{C}}$, with $\mathcal{C} = \{ i \in \mathbb{N} : I \le i \le I+M-1 \}$. The false-negative rate across the entire pixel population is computed as:
\begin{align}
\label{fnr}
\mathrm{FNR}_{\mathcal{D}_{\mathrm{cal}}}(\tau^{\mathrm{fail}}, \tau) = \frac{1}{N} 
\sum_{i \in \mathcal{C}} 
\sum_{\Omega'}  \mathbf{1}\{\mathbf{M}_i^{\tau^{\mathrm{fail}}} = 1 \wedge \hat{\mathbf{M}}_i^{\tau} = 0\},
\end{align}
\noindent where $N=\sum_{i \in \mathcal{C}} 
\sum_{\Omega'}  \mathbf{1}\{\mathbf{M}_i^{\tau^{\mathrm{fail}}} = 1 \}$ corresponds to the actual positive pixels, i.e., pixels with ground truth error above the target failure level. For a desired miscoverage, $\alpha$, and failure level, $\tau^{\mathrm{fail}}$, the operative threshold is:
\begin{align}
\label{threshold}
\tilde{\tau} = \sup \left\{ \tau :
\frac{N}{N+1} \,
\mathrm{FNR}_{\mathcal{D}_{\mathrm{cal}}}(\tau^{\mathrm{fail}}, \tau)
- \frac{1}{N+1}
\le \alpha \right\}.
\end{align}
\noindent Since $\mathrm{FNR}(\tau^{\mathrm{fail}}, \tau)$ is monotone in $\tau$, and analogously to standard conformal risk control proofs~\cite{angelopoulos2024conformal}, the bi-level finite-sample guarantee in Eq.~\ref{guarantee} will hold for new data that are, at least, exchangeable with the calibration distribution.

\section{Experiments}
\label{sec:experiments}

\subsection{Setup}
\label{ssec:setup}

\noindent\textbf{{Datasets}.} The proposed methods are validated on two different types of endoscopy data: robotic surgery imagery and minimally-invasive diagnostic videos. For the first scenario, we employ SurgiSR4K~\cite{jiang2025surgisr4k}, a super-resolution-focused dataset consisting of images at three resolution levels, up to native 4K. Two scenarios are contemplated: a 2$\times$ image enhancement, from 480x270p to 960x540p (\textbf{Surgi-2$\times$}), and a more challenging 4$\times$ super-resolution, from 960x540p up to 3840x2160p (\textbf{Surgi-4$\times$}). The samples are randomly split into training (10 videos, 335 images) and test (15 videos, 465 images) subsets. For the second scenario, HyperKvasir~\cite{borgli2020hyperkvasir} (\textbf{HKvasir}), which comprises endoscopic videos from gastrointestinal tract diagnostic procedures, is used. Specifically, we follow prior work~\cite{liu2025medvsr} and investigate a 4$\times$ SR enhancement with noise injection, from 180×144p to 720×576p images. Also, the same data splits as in~\cite{liu2025medvsr} are used. Training data in~\cite{liu2025medvsr} are used to pre-train a video SR model (236 videos, 11,800 images), while the validation (14 videos, 700 images) and test (27 videos, 1,350 images) subsets are used to train and evaluate our failure detection framework.

\noindent\textbf{{SR models}.} First, for the Surgi-s$\times$ datasets, we use SwinIR~\cite{liang2021swinir}, a transformer-based single-image SR model trained on natural images. Specifically, we use the default weights of its lightweight versions (\textbf{SwinIR-L-s$\times$}), which are designed for real-time SR applications (0.9M parameters; $\simeq$ 30 ms/image\footnote{Latency metrics are computed on a NVIDIA RTX A4000 16GB GPU workstation.}). Second, for HKvasir, we focus on video SR models, which propagate temporal features to improve consistency. Specifically, we used the lightweight \textbf{BasicVSR}~\cite{chan2021basicvsr} (2.9M parameters; $\simeq$ 20 ms/image). In our real-time scenario (only forward feature propagation allowed), we observed performance comparable to that of more recent methods, e.g., $\text{BasicVSR}^{++}$~\cite{chan2022basicvsr_pp} or MedVSR~\cite{liu2025medvsr}; therefore, we prioritized simplicity. Training hyper-parameters were as in the recent benchmark in~\cite{liu2025medvsr}.

\noindent\textbf{{Reconstruction Error Network}.} The network comprises 2 blocks, each with 64 features (70K parameters; latency negligible relative to SR). Using the small training split, as described in Section~\ref{ssec:failure_network}, a 30-epoch training run with a batch size of 1 video, the Adam optimizer, and a cosine-decay scheduler with a base learning rate of $10^{-3}$, is performed. All details remain fixed across tasks. 

\noindent\textbf{{Conformal Failure Masks (CFM)}} are created as described in Section~\ref{ssec:cfm}, at three target failure levels using cutoff ($\tau^{\mathrm{fail}}$) PSNR values of 22, 24, and 26~dB; and two FNR miscoverage, $\alpha\in\{0.10, 0.05\}$. Unless otherwise specified, exploratory experiments are conducted with a baseline failure level of 22~dB.

\noindent\textbf{{Metrics and evaluation protocol}.} Standard metrics for misclassification detection are used to assess the performance of our Reconstruction Error Network~\cite{corbiere2019addressing,granese2021doctor,lafon2025vilu}, i.e., area under the ROC curve (AUROC) and False Positive Rate at 95$\%$ True Positive Rate (FPR95). CFMs are assessed by measuring FNR at the target failure level. Also, the average mask size per image and PSNR in the non-rejected regions, as in \cite{adame2025image}, are obtained. The procedure is repeated and evaluated on $100$ random splits of the test videos ($70\%-30\%$ for $\mathcal{D}_{\mathrm{cal}}$ and test).

\subsection{Results}
\label{ssec:results}

\noindent\textbf{Main results.} Fig.~\ref{fig:failure_scores} confirms the positive outcome of our Reconstruction Error Network to discriminate erroneous reconstructions in the three use-cases, providing relatively low FNR95 results. We present the results of our constructed failure masks in Table~\ref {main_table}. First, it is worth noting that all SR models yield improved reconstructions, including SwinIR models pre-trained on natural datasets ($+2.2$ and $+0.7$ in PSNR on the SurgiSR datasets), compared to bicubic interpolation. Second, PSNR values significantly improve when considering only the non-rejected areas after introducing the CFMs ($\geq+2.2$, $\geq+0.9$, and $\geq+7.0$ for each dataset, respectively). The empirical miscoverage rate at each failure level closely aligns with the target $\alpha$ value while maintaining a reasonably small average rejected area, e.g., $4.0\%$ in Surgi-2$\times$ (${\tau}^{\mathrm{fail}}=22 \ \text{dB}$, $\alpha=0.05$), ensuring practical utility. Note that, for challenging domains such as 4K images or the noisy HKvasir, these masks are slightly increased to satisfy the desired risk level.

\begin{table}[t!]
\setlength{\tabcolsep}{3.5pt}
\centering
{\fontsize{8}{9}\selectfont
\caption{\textbf{Quantitative evaluation of our Conformal Failure Masks.}}
\label{main_table}
\begin{tabular}{llcccccc}

\toprule
& & \multicolumn{3}{c}{$\alpha=0.10$} &  \multicolumn{3}{c}{$\alpha=0.05$}   \\ \cmidrule(l){3-5}\cmidrule(l){6-8}
& & Avg. & FNR    & Avg. Mask   & Avg. & FNR    & Avg. Mask   \\
& SR Method & PSNR & ($\%$) & Size ($\%$) & PSNR & ($\%$) & size ($\%$) \\ \midrule
\multirow{5}{*}{\rotatebox{90}{\textbf{Surgi-2$\times$}}} & Bicubic  & 36.5 $\pm$ 1.2        & --          & --                    & 36.5 $\pm$ 1.2        & --          & --                    \\
& SwinIR-L-2$\times$                                                 & 38.7 $\pm$ 1.5        & --          & --                    & 38.7 $\pm$ 1.5        & --          & --                    \\
&  \grayc \hspace{1mm}+CFM (${\tau}^{\mathrm{fail}}=22 \ \text{dB}$) & \grayc 40.9 $\pm$ 0.8 & \grayc 10.8 & \grayc  2.7 $\pm$ 1.3 & \grayc 41.4 $\pm$ 0.7 &  \grayc 5.3 & \grayc  4.0 $\pm$ 1.7 \\  
&  \grayc \hspace{1mm}+CFM (${\tau}^{\mathrm{fail}}=24 \ \text{dB}$) & \grayc 41.3 $\pm$ 0.7 & \grayc 10.5 & \grayc  3.6 $\pm$ 1.6 & \grayc 41.8 $\pm$ 0.7 &  \grayc 5.1 & \grayc  5.4 $\pm$ 2.1 \\   
&  \grayc \hspace{1mm}+CFM (${\tau}^{\mathrm{fail}}=26 \ \text{dB}$) & \grayc 41.7 $\pm$ 0.7 & \grayc 10.4 & \grayc  4.8 $\pm$ 2.0 & \grayc 42.3 $\pm$ 0.7 &  \grayc 5.0 & \grayc  7.4 $\pm$ 3.1 \\     \midrule
\multirow{5}{*}{\rotatebox{90}{\textbf{Surgi-4$\times$}}} & Bicubic  & 33.0 $\pm$ 1.8        & --          & --                    & 33.0 $\pm$ 1.8        & --          & --                    \\
& SwinIR-L-4$\times$                                                 & 33.7 $\pm$ 1.9        & --          & --                    & 33.7 $\pm$ 1.9        & --          & --                    \\
&  \grayc \hspace{1mm}+CFM (${\tau}^{\mathrm{fail}}=22 \ \text{dB}$) & \grayc 34.6 $\pm$ 1.2 & \grayc  9.6 & \grayc  4.0 $\pm$ 4.2 & \grayc 34.7 $\pm$ 1.2 &  \grayc 4.6 & \grayc  5.4 $\pm$ 5.2 \\  
&  \grayc \hspace{1mm}+CFM (${\tau}^{\mathrm{fail}}=24 \ \text{dB}$) & \grayc 34.7 $\pm$ 1.2 & \grayc  9.5 & \grayc  5.0 $\pm$ 4.9 & \grayc 34.8 $\pm$ 1.2 &  \grayc 4.5 & \grayc  6.9 $\pm$ 6.0 \\   
&  \grayc \hspace{1mm}+CFM (${\tau}^{\mathrm{fail}}=26 \ \text{dB}$) & \grayc 34.9 $\pm$ 1.2 & \grayc  9.3 & \grayc  8.9 $\pm$ 6.6 & \grayc 35.1 $\pm$ 1.2 &  \grayc 5.1 & \grayc 17.7 $\pm$ 9.7 \\     \midrule
\multirow{5}{*}{\rotatebox{90}{\textbf{HKvasir}}} & Bicubic          & 26.0 $\pm$ 0.7        & --          & --                    & 26.0 $\pm$ 0.7        & --          & --                    \\
& BasicVSR                                                           & 30.4 $\pm$ 1.2        & --          & --                    & 30.4 $\pm$ 1.2        & --          & --                    \\
&  \grayc \hspace{1mm}+CFM (${\tau}^{\mathrm{fail}}=22 \ \text{dB}$) & \grayc 37.4 $\pm$ 2.1 & \grayc 10.2 & \grayc  5.6 $\pm$ 2.6 & \grayc 38.3 $\pm$ 2.1 &  \grayc 5.0 & \grayc 10.0 $\pm$ 4.3 \\  
&  \grayc \hspace{1mm}+CFM (${\tau}^{\mathrm{fail}}=24 \ \text{dB}$) & \grayc 37.9 $\pm$ 2.1 & \grayc 10.2 & \grayc  7.5 $\pm$ 3.4 & \grayc 38.7 $\pm$ 2.0 &  \grayc 5.0 & \grayc 13.7 $\pm$ 5.3 \\   
&  \grayc \hspace{1mm}+CFM (${\tau}^{\mathrm{fail}}=26 \ \text{dB}$) & \grayc 38.3 $\pm$ 2.1 & \grayc 10.2 & \grayc 10.3 $\pm$ 4.4 & \grayc 39.2 $\pm$ 2.0 &  \grayc 5.0 & \grayc 19.4 $\pm$ 6.3 \\ 
\bottomrule

\end{tabular}
}
\end{table}
\begin{figure*}[t!]
\setlength{\tabcolsep}{4.0pt}
    \begin{center}
        \begin{tabular}{ccc}

        \includegraphics[width=.30\linewidth]{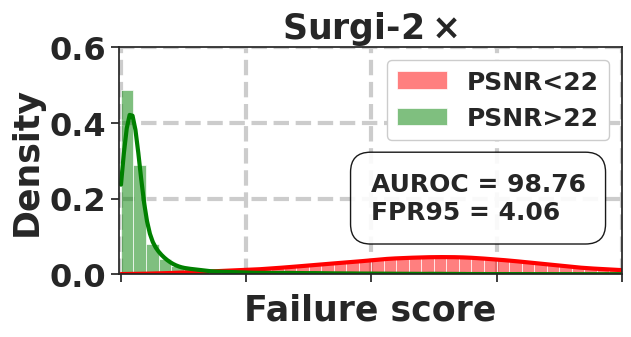} &
        \includegraphics[width=.30\linewidth]{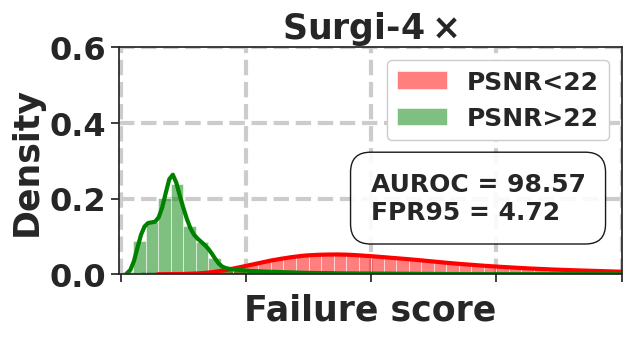} &
        \includegraphics[width=.30\linewidth]{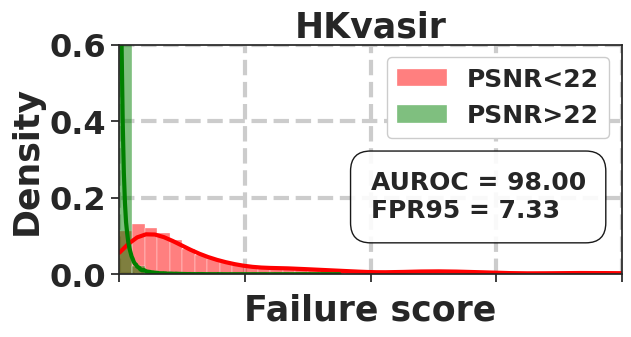} \\

        \end{tabular}
        \caption{\textbf{Reconstruction Error Network' scores performance}.}
        \label{fig:failure_scores}
    \end{center}
\end{figure*}

\begin{figure*}[t!]
\setlength{\tabcolsep}{4.0pt}
    \begin{center}
        \begin{tabular}{ccc}

        \includegraphics[width=.33\linewidth]{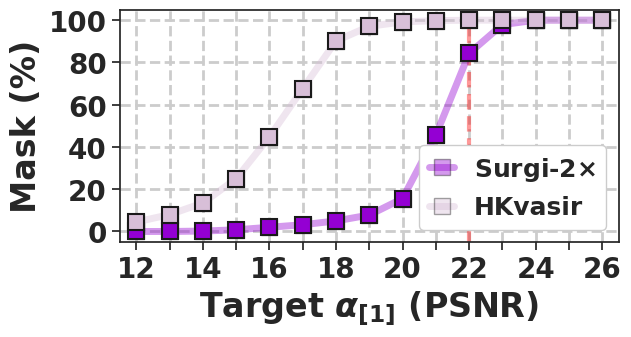} &
        \includegraphics[width=.33\linewidth]{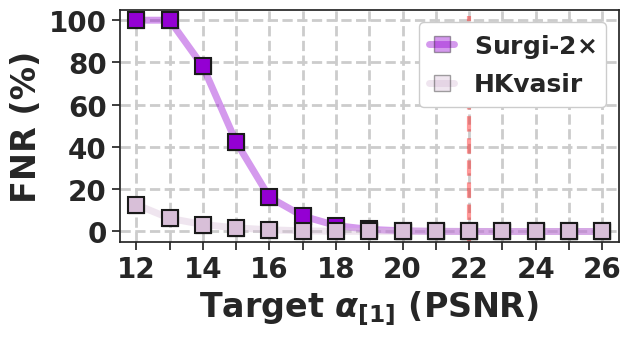} &
        \includegraphics[width=.27\linewidth]{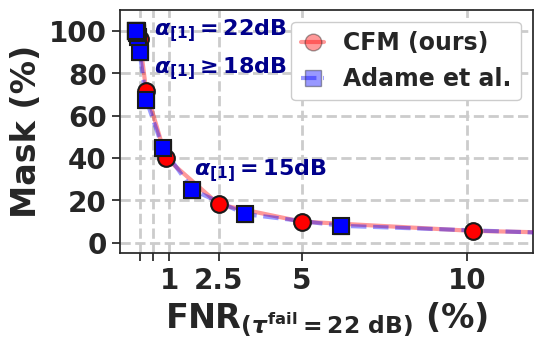} \\

        \multicolumn{2}{c}{\textbf{(a)} Conservativeness and inconsistency across datasets} & \textbf{(b)} FNR control

        \end{tabular}
        \caption{\textbf{Comparison of our CFM with Adame \textit{et al.} \cite{adame2025image} procedure}.}
        \label{fig:conf_method}
    \end{center}
\end{figure*}

\noindent\textbf{Comparison with Adame \textit{et al.}~\cite{adame2025image}.} This recent effort to construct SR rejection masks with guarantees focuses on controlling the maximum error per image. In~\cite{adame2025image}, the meaning of $\alpha$ differs: it does not control miscoverage but rather the allowed peak error; e.g., $\alpha_{[1]}$ denotes the minimum PSNR among the non-rejected pixels. In practice, this results in successful, but overly conservative masks that cover almost the entire image (\textit{see} Fig.~\ref{fig:conf_method}(a), $\alpha_{[1]}=22 \ \text{dB}$). Even though $\alpha_{[1]}$ can be adjusted to reduce the mask size, the procedure in~\cite{adame2025image} lacks direct control over miscoverage at each desired failure level, as this trade-off is task-specific. In contrast, our CFMs simultaneously control the desired failure level and FNR, making them a more flexible solution. We aim to exemplify such a lack of control in Fig.~\ref{fig:conf_method}(b) on the HKvasir dataset. Concretely, we plot the FNR for a target failure level of ${\tau}^{\mathrm{fail}}=22 \ \text{dB}$ against the average mask size of different operative thresholds with $\alpha \ \text{(\%)}=\{0.1, 0.5, 1, 2.5, 5, 10\}$ for our method (\textit{red dots}), against $\alpha_{[1]} \ \text{(dB)}=\{13, ..., 22, ..., 26\}$ (\textit{blue squares}) for the procedure in~\cite{adame2025image}. As previously pointed out,~\cite{adame2025image} does not produce almost any false negatives, but at the cost of creating large masks when applied at the target PSNR level, i.e., $\alpha_{[1]}=22 \ \text{dB}$. For example, if one would require more relaxed masks for usability purposes, e.g., covering $\leq20\%$ of the image on average, then one should set $\alpha_{[1]}\sim15 \ \text{dB}$ or below, which is quite arbitrary, and is not linked with any particular FNR level. In contrast, for our bi-level procedure, as per design in Eq.~\ref{guarantee}, FNR is successfully controlled by $\alpha$ at each ${\tau}^{\mathrm{fail}}$. Indeed, if more conservative masks were desired, the user would only need to decrease $\alpha$, as shown in Fig.~\ref{fig:conf_method}(b). Hence, the conformal procedure in~\cite{adame2025image} can be seen as a specific, conservative case of our general framework, where we set $\alpha \rightarrow 0$ for a given ${\tau}^{\mathrm{fail}}$.

\begin{figure*}[t!]
\setlength{\tabcolsep}{4.0pt}
    \begin{center}
        \begin{tabular}{ccc}

        \includegraphics[width=.30\linewidth]{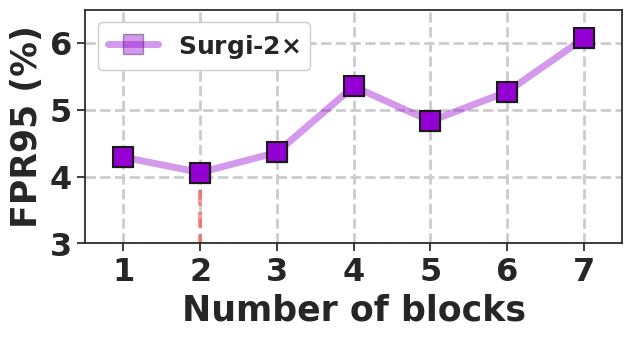} &
        \includegraphics[width=.30\linewidth]{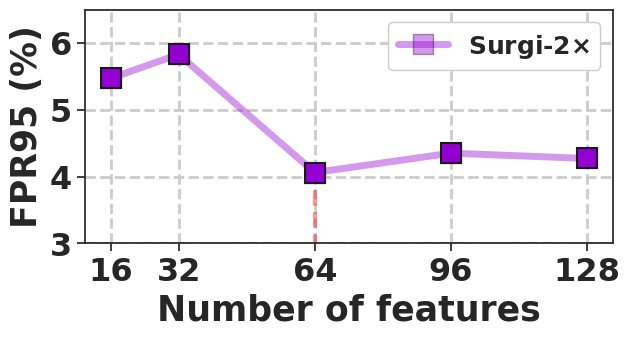} &
        \includegraphics[width=.30\linewidth]{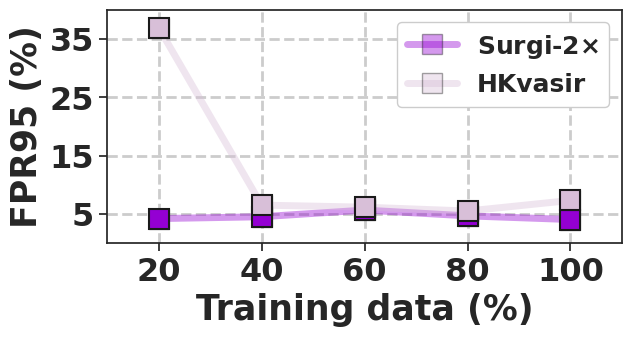} \\

        \end{tabular}
        \caption{\textbf{Reconstruction Error Network configuration and data-efficiency}.}
        \label{fig:ablation_studies}
    \end{center}
\end{figure*}

\noindent\textbf{Network configuration.} Fig.~\ref{fig:ablation_studies} demonstrates effectiveness with reduced complexity (\textit{left} \& \textit{center}), and promising data efficiency (\textit{right}), requiring only a few videos for optimal performance. Note also that this module was trained on HKvasir with videos already used for validation during SR model pre-training (e.g., hyper-parameter tuning), indicating successful reuse of this data.

\noindent\textbf{CFMs visualizations.} Fig.~\ref{fig:qualitative} (\textit{best zoomed-in}) shows the produced CFMs for Surgi-4$\times$ (\textit{top}) and HKvasir (\textit{bottom}). These typically detect challenging lighting conditions, e.g., reflections (B), incorrectly reconstructed blood vessels (A), smoke artifacts along borders (C), and overly smoothed patterns (G).

\section{Discussion}
\label{sec:conclusion}

This work demonstrates the potential of our CFMs to localize erroneous regions in endoscopic SR. While they guarantee control over false negatives, their quality depends on accurate error prediction, as poor estimates can yield overly conservative masks; improving these predictions under real-time constraints remains future work. Our guarantees inherit known limitations of conformal methods in medical imaging~\cite{Mehrtens2025PitfallsOC}, including reliance on calibration size, marginal coverage, and exchangeability assumptions. Although calibration size is less restrictive at the pixel level, the latter assumptions warrant further investigation~\cite{lofstrom2015bias,ding2023classconditional,confbeyondEx_jour,confbeyondEx_conf}.

\begin{credits}
\subsubsection{\ackname} Innovation project supported by Innosuisse (122.308 IP-LS).
\end{credits}

\bibliographystyle{splncs04}
\bibliography{refs}

\end{document}